\documentclass[sigconf]{aamas}  
\usepackage{balance}  

\usepackage{booktabs}

\setcopyright{ifaamas}  
\acmDOI{}  
\acmISBN{}  
\acmConference[AAMAS'19]{Proc.\@ of the 18th International Conference on Autonomous Agents and Multiagent Systems (AAMAS 2019)}{May 13--17, 2019}{Montreal, Canada}{N.~Agmon, M.~E.~Taylor, E.~Elkind, M.~Veloso (eds.)}  
\acmYear{2019}  
\copyrightyear{2019}  
\acmPrice{}  

\begin{document}

\title{A Grounded Interaction Protocol for Explainable Artificial Intelligence}  


\author{Prashan Madumal}
\affiliation{%
  \institution{University of Melbourne}
  \city{Victoria, Australia} 
}
\email{pmathugama@student.unimelb.edu.au}
\author{Tim Miller}
\affiliation{%
  \institution{University of Melbourne}
  \city{Victoria, Australia} 
}
\email{tmiller@unimelb.edu.au}

\author{Liz Sonenberg}
\affiliation{%
  \institution{University of Melbourne}
  \city{Victoria, Australia} 
}
\email{l.sonenberg@unimelb.edu.au}

\author{Frank Vetere}
\affiliation{%
  \institution{University of Melbourne}
  \city{Victoria, Australia} 
}
\email{f.vetere@unimelb.edu.au}

\begin{abstract}  
Explainable Artificial Intelligence (XAI) systems need to include an explanation model to communicate the internal decisions, behaviours and actions to the interacting humans. Successful explanation involves both cognitive and social processes. In this paper we focus on the challenge of meaningful interaction between an explainer and an explainee and investigate the structural aspects of an interactive explanation to propose an interaction protocol. We follow a bottom-up approach to derive the model by analysing transcripts of different explanation dialogue types with \textbf{398} explanation dialogues. We use grounded theory to code and identify key components of an explanation dialogue. We formalize the model using the agent dialogue framework (ADF) as a new dialogue type and then evaluate it in a human-agent interaction study with \textbf{101} dialogues from \textbf{14} participants. Our results show that the proposed model can closely follow the explanation dialogues of human-agent conversations.
\end{abstract}

%


\maketitle


\section{Introduction}

In scenarios where people are required to make critical choices based on decisions from an artificial intelligence (AI) system, it is important for the system to able to generate understandable explanations that clearly justify its decisions. An appropriate explanation can promote trust in the system, allowing better human-AI cooperation~\cite{Reference16}. Explanations also help people to reason about the extent to which, if at all, they should trust the provider of the explanation.

	As Miller~\shortcite[pg 10]{Reference13} notes, the process of Explanation involves two processes: (a) a \emph{Cognitive process}, namely the process of determining an explanation for a given event, called the \emph{explanandum}, in which the causes for the event are identified and a subset of these causes is selected as the explanation (or \emph{explanans}); and (b) the \emph{Social process} of transferring knowledge between explainer and explainee, generally an interaction between a group of people, in which the goal is that the explainee has enough information to understand the causes of the event.
    
However, much research and practice in explainable AI uses the researchers' intuitions of what constitutes a `good' explanation rather basing the approach on a strong understanding of how people define, generate, select, evaluate, and present explanations~\cite{Reference13,miller2017explainable}. Most modern work on Explainable AI, such as in autonomous agents~\cite{winikoff2017debugging,broekens2010you,Reference20,Reference21} and interpretable machine learning~\cite{Reference19}, does not discuss the interaction and the social aspect of the explanations. The lack of a general interaction model of explanation that takes into account the end user can be attributed as one of the shortcomings of existing explainable AI systems. Although there are existing conceptual explanation dialogue models that try to emulate the structure and sequence of a natural explanation~\cite{Reference6,Reference4}, we propose that improvements will come from further empirically-driven study of explanation.

Explanation naturally occurs as a continuous interaction, which gives the interacting party the ability to question and interrogate explanations. This allows the explainee to clear doubts about the given explanation by further interrogations and user-driven questions. Further, the explainee can express contrasting views about the explanation that can set the premise for an argumentation based interaction. This type of iterative explanation can provide richer and satisfactory explanations as opposed to one-shot explanations. Note that we are not claiming that AI explanations are necessarily textual conversations. These interactions, questions, and answers can occur as part of other modalities, such as visualisations, but we believe that such interactions will follow the same model.

Understanding how humans engage in conversational explanation is a prerequisite to building an explanation model, as noted by Hilton~\shortcite{Reference22}. De Graaf~\shortcite{Reference3} note that humans attribute human traits, such as beliefs, desires, and intentions, to intelligent agents, and it is thus a small step to assume that people will seek to explain agent behaviour using human frameworks of explanation. We hypothesise that AI explanation models with designs that are influenced by human explanation models have the potential to provide more intuitive explanations to humans and therefore be more likely to be understood and accepted. We suggest it is easier for the AI to emulate human explanations rather than expecting humans to adapt to a novel and unfamiliar explanation model. While there are mature existing models for explanation dialogs~\cite{Reference4,Reference15}, these are idealised conceptual models that are not grounded on or validated by data, and seem to lack iterative features like cyclic dialogues.

In this paper our goal is to introduce a dialogue model and an interaction protocol that is based on data obtained from different types of explanations in actual conversations. We derive our model by analysing 398 explanation dialogues using grounded theory~\cite{Reference9} across six different dialogue types. Frequency, sequence and relationships between the basic components of an explanation dialogue were obtained and analyzed in the study to identify locutions, termination rules and combination rules. We formalize the explanation dialogue model using the \emph{agent dialogue framework} (ADF)~\cite{McBurney2002}, then validate the model in a human-agent study with 101 explanation dialogues. We propose that by following a data-driven approach to formulate and validate, our model more accurately defines the structure and the sequence of an explanation dialogue and will support more natural interaction with human audiences than explanations from existing models. The main contribution of this paper is a grounded interaction protocol derived from explanation dialogues, formalized as a new atomic dialogue type~\cite{Walton1995-WALCID} in the ADF.

We first discuss related work regarding explanation in AI and explanation dialogue models, then we outline the methodology of the study and collection of data and its properties. We then present the analysis of the data, identifying key components of an explanation dialogue and gaining insight to the relationships of these components, formalising it using ADF and comparing with a similar conceptual model~\cite{Reference8}. We then describe the human-agent study and present the validation of the model. 
We conclude by discussing the model with its contribution and significance in explainable AI.

\section{Related Work}
Explaining decisions of intelligent systems has been a topic of interest since the era of expert systems, e.g.~\cite{Reference1,KassFinin88}. Early work focused particularly on the explanation's content, responsiveness and the human-computer interface through which the explanation was delivered. 
Kass and Finin~\shortcite{KassFinin88} and Moore and Paris~\shortcite{Reference17} discussed the requirements a good explanation facility should have, 
including characteristics like ``Naturalness'', and pointed to the critical role of user models in explanation generation.
Cawsey's~\shortcite{Reference2} EDGE system also focused on user interaction and user knowledge. These were used to update the system through interaction. So, in early explainable AI, both the  cognitive and social attributes associated with an agent's awareness of other actors, and capability to interaction with them, has been recognized as an essential feature of explanation research. However, limited progress has been made. Indeed recently, de Graaf and Malle~\shortcite{Reference3} still find the need to emphasize the importance of understanding how humans respond to Autonomous Intelligent Systems (AIS). They further note how humans will expect a familiar way of communication from AIS systems when providing explanations.


To accommodate the communication aspects of explanations, several dialogue models have been proposed.  Walton~\shortcite{Reference4,Reference15} introduces a shift model that has two distinct dialogues: an explanation dialogue and an examination dialogue, where the latter is used to evaluate the success of an explanation. Walton draws from the work of Memory Organizing Packages (MOP)~\cite{Reference18} and case-based reasoning to build the routines of the explanation dialogue models. Walton's dialogue model has three stages: opening, argumentation, and closing~\cite{Reference4}. Walton suggests an examination dialogue with two rules as the closing stage. These rules are governed by the explainee, which corresponds to the understanding of an explanation~\cite{Reference5}. This sets the premise for the examination dialogue of an explanation and the shift between explanation and examination to determine the success of an explanation~\cite{Reference15}. 

A formal dialogical system of explanation is also proposed by Walton~\shortcite{Reference5}. This has three types of conditions: dialogue conditions, understanding conditions, and success conditions. Arioua~\cite{Reference6} formalize and extend Walton's dialectical system by incorporating Prakken's~\shortcite{Reference7} framework of dialogue formalisation.

Argumentation also comes into to play in explanation dialogues. Walton and Bex~\shortcite{Reference8} introduce  a dialogue system for argumentation and explanation that consists of a communication language that defines the speech acts and protocols that allow transitions in the dialogue. This allows the explainee to challenge and interrogate the given explanations to gain further understanding. ~\citet{Reference16} focus on modelling information sources to be suited in an argumentation framework, and introduce a socio-cognitive model of trust to support judgements about trustworthiness. 

This previous work on explanation dialogues is largely conceptual and involves idealized models, and mostly lacks empirical validation. In contrast, we take a grounded, data-driven approach to determine what an explanation dialogue should look like.

\section{Methodology}

To address the lack of a grounded explanation interaction protocol, we studied real conversational data of explanations. This study consists of data selection and gathering, data analysis, and model development, and then a validation in a lab based simulated human-agent experiment.

We designed a bottom-up study to develop an explanation dialogue model. We aimed to gain insights into three areas: 1. key components that makeup an explanation interaction protocol (locutions); 2. relationships within those components (termination rules); and 3. component sequences and cycles (combination rules) that occur in explanations. 

\subsection{Design}
We formulate our design based on an inductive approach. We use grounded theory~\cite{Reference9} as the methodology to conceptualize and derive models of explanation. The key goal of using grounded theory, as opposed to using a hypothetico-deductive approach, is to formalize a model that is grounded on actual conversation data of various types, rather than a purely conceptual model.

The study is divided into three distinct stages, based on grounded theory. The first stage consists of coding~\cite{Reference9} and theorizing, where small chunks of data are taken, named and marked \emph{manually} according to the concepts they might hold. For example, a segment of a paragraph in an interview transcript can be identified as an `Explanation' and another segment can be identified as a `Why question'. This process is repeated until the whole data set is coded. The second stage is categorizing, where similar codes and concepts are grouped together by identifying their relationship with each other. The third stage derives a theoretical model from the codes, categories and their relationship.

\subsection{Data}

We collected data from six different data sources encompassing six different types of explanation dialogues. Table~\ref{tab1} shows the explanation dialogue types, explanation dialogues that are in each type and number of transcripts. Here, `static' is defined as when an explainee or an explainer is the same person change from transcript to transcript (e.g.\ same journalist interviewing different people). We gathered and coded a total of 398 explanation dialogues from all of the data sources. All the data sources\footnote{\label{note1}Links to all  data sources (including transcripts) can be found at \url{https://explanationdialogs.azurewebsites.net}} are text based, where some of them are transcribed from voice and video-based interviews. Data sources consist of Human-Human conversations and Human-Agent conversations. We collected Human-Agent conversations to analyze if there are significant differences in the way humans carry out the explanation dialogue when they knew the interacting party was an agent with respect to the frequency of different locutions. 

\begin{table}[!h]
\caption{Coded data description.}
\label{tab1}

\begin{center}
\begin{small}

\begin{tabular}{lrr}
\toprule
Explanation Dialogue Type &  \#Dialogue & ~~\#Scripts \\
\midrule
1. Human-Human static explainee &  88 & 2 \\
2. Human-Human static explainer & 30 & 3 \\
3. Human-Explainer agent & 68 & 4 \\
4. Human-Explainee agent & 17 & 1 \\
5. Human-Human QnA & 50  & 5 \\
6. Human-Human multiple explainee & 145 & 5 \\
\bottomrule
\end{tabular}

\end{small}
\end{center}

\end{table}

Data source selection was done to encompass different combinations of participant types and numbers. These combinations are given in Table \ref{tab3}. We diversify the dataset by including data sources of different mediums such as verbal based and text based.

\begin{table}[!h]
\caption{Explanation dialogue type description.}
\label{tab3}
\begin{center}
\begin{small}

\begin{tabular}{p{2.6cm}p{1cm}p{1cm}p{2cm}}
\toprule
Participants &  Number & Medium~ & Data source  \\
\midrule
1. Human-Human~  &  1-1 & Verbal & Journalist Interview transcripts \\
2. Human-Human  & 1-1 & Verbal & Journalist Interview transcripts \\
3. Human-Agent & 1-1 & Text & Chatbot conversation transcripts \\
4. Human-Agent & 1-1 & Text & Chatbot conversation transcripts \\
5. Human-Human & n-m~ & Text  & Reddit AMA records \\
6. Human-Human & 1-n & Verbal & Supreme court transcripts \\
\bottomrule
\end{tabular}
\end{small}
\end{center}
\end{table}

Table \ref{tab2} presents the codes and their definitions. We identify `why', `how' and `what' questions as questions that ask counterfactual explanations, questions that ask explanations of causal chains, and questions that ask causality explanations respectively. The whole number of the code column refers to the categories the codes belong to, where 1) Dialogue boundary; 2) Question type; 3) Explanation; 4) Argumentation; 5) Return question type.

\begin{table}[!h]
\caption{Code description.}
\label{tab2}

\begin{center}
\begin{small}

\begin{tabular}{p{3cm}p{4.5cm}}
\toprule
Code  & Description  \\
\midrule
1.1 QE start & Explanation dialogue start\\
1.2 QE end  & Explanation dialogue end\\
2.1 How  & How questions\\
2.2 Why  & Why questions\\
2.3 What   & What questions \\
3.1 Explanation  & Explanation given for questions\\
3.2 Explainee Affirmation & Explainee acknowledges explanation 
 \\
3.3 Explainer Affirmation &  Explainer acknowledges explainee's acknowledgment\\
3.4 Question context & Background to the question provided by the explainee\\
3.5 Counterfactual case & Counterfactual case of the how/why question  \\
4.1 Argument & Argument presented by explainee or explainer  \\
4.2 Argument-s  & An argument that starts the dialogue\\
4.3 Argument-a & Argument Affirmation by explainee or explainer\\
4.4 Argument-c  & Counter argument\\
4.5 Argument-contrast case  & Argumentation contrast case
\\
5.1 Explainer Return question & Clarification question by explainer\\
5.2 Explainee Return question  & Follow up question asked by explainee
 \\
\bottomrule
\end{tabular}

\end{small}
\end{center}

\end{table}

\section{Grounded Explanation Interaction Protocol}

\begin{figure*}[!ht]
\includegraphics[width=\textwidth]{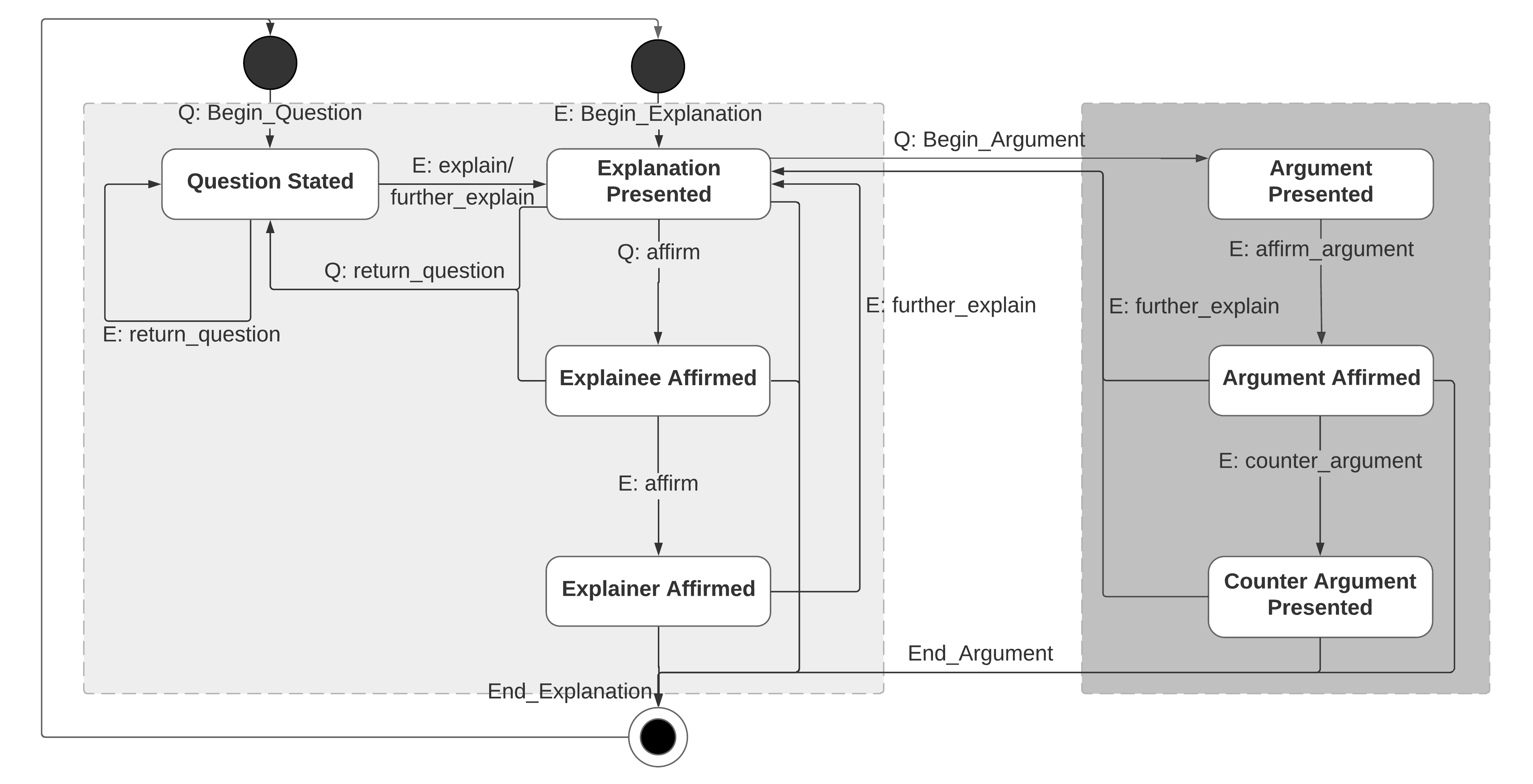}
\caption{Explanation Dialogue Model} \label{fig5}
\end{figure*}

In this section, we present the interaction model resulting from our grounded study, and formalize the model using agent dialogue framework (ADF) \cite{McBurney2002} as an atomic dialogue type~\cite{Walton1995-WALCID}. Note that a dialogue can range from a purely visual user interface interaction to verbal interactions. We analyse some observed patterns of interaction and  compare the grounded model to an existing conceptual model.  

When formalizing the model, we consider the interaction between explainer and explainee as a dialogue game. Dialogue games are depicted as interactions between two or more players. The players can make `moves' with utterances, according to a set of rules. Dialogue game models have been used to model human-computer interaction \cite{Dunne2014InteractingWK}, to model human reasoning \cite{Prakken1998} and to develop protocols for interactions between agents \cite{Dignum}.

Formal dialogue models have been proposed for different dialogue types \cite{Walton1995-WALCID}, such as negotiation dialogues \cite{Amgoud}, persuasion dialogues \cite{Walton1995-WALCID} and a combination of negotiation and persuasion dialogues \cite{Dignum}. To the best of our knowledge there is no formal explanation dialogue game model grounded on data.

\subsection{Agent Dialogue Framework}

We use McBurney and Parson's agent dialogue framework \cite{McBurney2002} to formalize the explanation dialogue model as a dialogue game. The Agent Dialogue Framework (ADF) provides a modular and unifying framework that can represent and combine different types of atomic dialogues in the typology of \citeauthor{Walton1995-WALCID} \shortcite{Walton1995-WALCID}, with the freedom of introducing new dialogue type combinations. The ADF has three layers: 1. topic layer; 2. dialogue layer; and 3. control layer. In the topic layer, the topics of discussion in a dialogue game are presented in a logical language. Then, the dialogue layer \cite{McBurney2002} consists of a set of rules:

{\textbf{Commencement rules:}} rules under which the dialogue commences.

{\textbf{Locutions:}} Rules that determine which utterances are permitted in the dialogue-game. Typical locutions include assertions, questions, arguments, etc.

{\textbf{Combination rules:}} Rules that define the dialogical context of the applicability of locutions. E.g.\ it might not be applicable to assert preposition $p$ and $\neg p$ in the same dialogue.

{\textbf{Commitments:}} Rules that determine the circumstances where players express commitments to a preposition.

{\textbf{Termination rules:}} Rules that determine the ending of a dialogue.

More formally, given a set of participating agents $\mathcal{A}$, we define the dialogue $G$ at the dialogue layer as a 4-tuple $\left ( \Theta, \mathcal{R}, \mathcal{T}, \mathcal{CF}  \right )$, where $\Theta$ denotes set of legal locutions, $\mathcal{R}$ the set of combinations, $\mathcal{T}$ the set of termination rules and $\mathcal{CF}$ the set of commitment functions respectively \cite{McBurney2002}. 

Selection and transitions between dialogue types are handled in the control layer. Dialogue types can be combined using iteration, sequencing and embedding \cite[pg 10]{McBurney2002}. 
When combined, an ADF is given by 5-tuple $\left (\mathcal{A}, \mathcal{L}, \Pi_{a}, \Pi_{c}, \Pi  \right )$ where the set of agents is given by $\mathcal{A}$, logical language representation given by $\mathcal{L}$, set of atomic dialogue types given by $\Pi_{a}$, set of control dialogues given by $\Pi_{c}$, and $\Pi$ is the closure of  $\Pi_a\cup\Pi_c$, which represents the set of formal dialogues denoted by the 4-tuple given above. Closure is defined under the combination rules presented by McBurney and Parsons \shortcite{McBurney2002}.

\subsection{Formal Explanation Dialogue Game Model}

In this section we present the formal explanation dialogue model as a new atomic dialogue type~\cite{Walton1995-WALCID} using the modular ADF, and discuss how it is derived from the grounded data of explanation dialogues according to the layers of ADF. Our analysis of the data shows that people switch from explanation to argumentation and back again during an explanation dialogue, in which the explainee questions a claim made by an explainer. For this reason, our model has two dialogue types: Explanation and Argumentation. Dialogue games of atomic dialogue types~\cite{Walton1995-WALCID} have an initial situation and an aim (e.g persuasion dialogue having the initial situation of conflicting opinions of the interacting party and the aim of resolving the conflict). For our explanation dialogue, the initial condition is the knowledge discrepancy between explainer and explainee of the topic $p$ and the aim is to provide knowledge about the topic $p$ to the explainee.

Formally, the explanation dialogue model ($ADF_E$) is the tuple: 
\vspace{-1ex}
\begin{equation} \label{eq:1}
ADF_E = \left (\mathcal{A}, \mathcal{L}, \Pi_{a}, \Pi_{c}, \Pi  \right )
\end{equation}
where the set of agents $\mathcal{A} = \{Q, E\}$, where labels $Q$ and $E$ refer to the Questioner (the explainee) and the Explainer respectively; $\mathcal{L}$ is the set of logical representations about topics (denoted by $p$, $q$, $r$, ...), $\Pi_{a} = \{G_E, G_A\}$, where $G_E$ is the explanation dialogue and $G_A$ is the argumentation dialogue, $\Pi_{c} =$ (Begin\_Question, Begin\_Explanation, Begin\_Argument, End\_Explanation, End\_Argument), and $\Pi$ is the closure of $\Pi_a \cup \Pi_c$ under the combination rule set. $\Pi$ gives us the set of formal explanation dialogue $G$.

The \textbf{Topic Layer} is dependent on the particular application domain in which the explanation dialogue is embedded, so we do not define this further.

\paragraph{Dialogue Layer} The dialogue layer consists of the two dialogue types: explanation ($G_E$) and argumentation ($G_A$):
\begin{equation} \label{eq:2}
\begin{split}
& G_{E} = \left ( \Theta_E, \mathcal{R}_E, \mathcal{T}_E, \mathcal{CF}_E  \right ) \\
& G_{A} = \left ( \Theta_A, \mathcal{R}_A, \mathcal{T}_A, \mathcal{CF}_A  \right )
\end{split}
\end{equation}
%
The set of legal locutions are defined by:
\begin{equation}
\begin{split}
& \Theta_E = (explain, \textit{affirm}, \textit{further}\_explain, return\_question)\\
& \Theta_A = (\textit{affirm}\_argument, counter\_argument, \textit{further}\_explain).   
\end{split}
\end{equation}
For clarity, we define the commencement rules, combination rules, and termination rules via the state transition diagram in Figure~\ref{fig5}. While most codes are directly transferred to the model as states and state transitions, codes that belonged to information category are embedded in different states.  The combination rules $\mathcal{R}_E$ and $\mathcal{R}_A$ are defined by the individual transitions on the diagram. For example, after a dialogue begins with a question, the next locution is either the explainer asking for clarification using a return\_question or giving an explanation. Similarly, the set of termination rules can be extracted from the state model as the state transitions that lead to the termination state, giving  $\mathcal{T}_E =$ (affirm(p), explain(p)) and $\mathcal{T}_A =$ (affirm\_argument(p), counter\_argument(p)). We do not define commitments $\mathcal{CF}$ as these were not observable in our data. 

\paragraph{Control layer} This can be identified as state transitions that lead to and out of the two dialogue types in Figure~\ref{fig5} (e.g.\ argue, explanation\_end). Argumentation occurs naturally within explanation dialogues, meaning that this is an \emph{embedded} dialogue, as defined by \citeauthor{McBurney2002} \shortcite{McBurney2002}. An argument can occur after an explanation was given, which will then continue on to an argumentation dialogue. The dialogue then returns to the explanation dialogue, as shown in Figure \ref{fig5}. A single explanation dialogue can contain many embedded argumentation dialogues. 

Explanation dialogues can occur in sequence, which is modelled by the external loop. Note that a loop within the explanation dialogue implies that the ongoing explanation is related to the same original question and topic, while a loop outside of the dialogue means a new topic is introduced. We coded explanation dialogues to end when a new topic was raised in a question. Questions that ask for follow-up explanations (return\_question) were coded when the questions were clearly identifiable as requesting more information about the given explanation.  

\textbf{Example:} We now go through the formal model with an example dialogue which is taken from the human-agent experiments discussed in Section 5.1. Example is given in Table \ref{tab4} with the dialogue text, locutions/rules and a  commentary about the dialogue. Two agents who are explainee (player) and the explainer (agent) participate in the dialogue given by $Q$ and $E$ respectively and the topic `cities' by $p$:


\begin{table}[!h]
\caption{Example: from human-agent experiments of Ticket to Ride domain.}
\label{tab4}

\begin{center}
\begin{small}

\begin{tabular}{p{3cm}p{2.2cm}p{2.2cm}}
\toprule
Dialogue Text &  Locutions/Rules & Commentary \\
\midrule
\textbf{$E$: Opponent is gazing at Duluth to Omaha route and will try to extend it to Kansas City.} &  \textit{Begin\_Explanation(p)} & - Commence explanation dialogue with an explanation about cities which the opponent is gazing at $p$. \\
\textbf{$Q$: Is he going to Pittsburgh?} & \textit{return\_question(p)} & - Using locution \textit{return\_question} available in Explanation dialogue type, inquiring more information \\
\textbf{$E$: Opponent will try to Extend the path from Pittsburgh to Houston through Atlanta, has been repeatedly gazing at that path} & \textit{further\_explain(p)} & - providing further explanation using \textit{further\_explain} locution about topic p. \\
\textbf{$Q$: No. He is going to El Paso.} & \textit{Begin\_Argument(p)} & - Argumentation sub-dilaog begins about topic $p$ after. \\
\textbf{$E$: Yes, now opponents gaze is focused at El Paso and will try to build from Little rock to Dallas to El Paso.} & \textit{affirm\_argument(p)}  & - Argument is acknowledged by the Agent (E) using locution \textit{affirm\_argument} \\
. & \textit{End\_Argument(p)} & - End the embedded argument dialogue by using the control dialogue \textit{End\_Argument} which also ends the initial dialogue.\\
\bottomrule
\end{tabular}

\end{small}
\end{center}

\end{table}

This example shows an interaction between an agent and a human using the explanation dialogue with an embedded argumentation dialogue. The human-agent study is discussed in depth in Section 5.1. The example demonstrates the ability of our model to handle embedded dialogues and cyclic dialogues (explanation dialogues that occurs twice with Begin\_explanation and further\_explain) which similar model of explanation dialogue by Walton~\cite{Reference8} lack. A detailed model comparison between our model and Walton's can be found in Section 4.4.

\subsection{Analysis}

We focus our analysis on three areas to further reinforce the derived interaction protocol: 1. Key components of an Explanation Dialogue; 2. Relationships between these components and their variations between different dialogue types; and 3. The sequence of components that can successfully carry out an explanation dialogue.

\subsubsection{Code Frequency Analysis}

\begin{figure}[!h]
\includegraphics[width=\columnwidth]{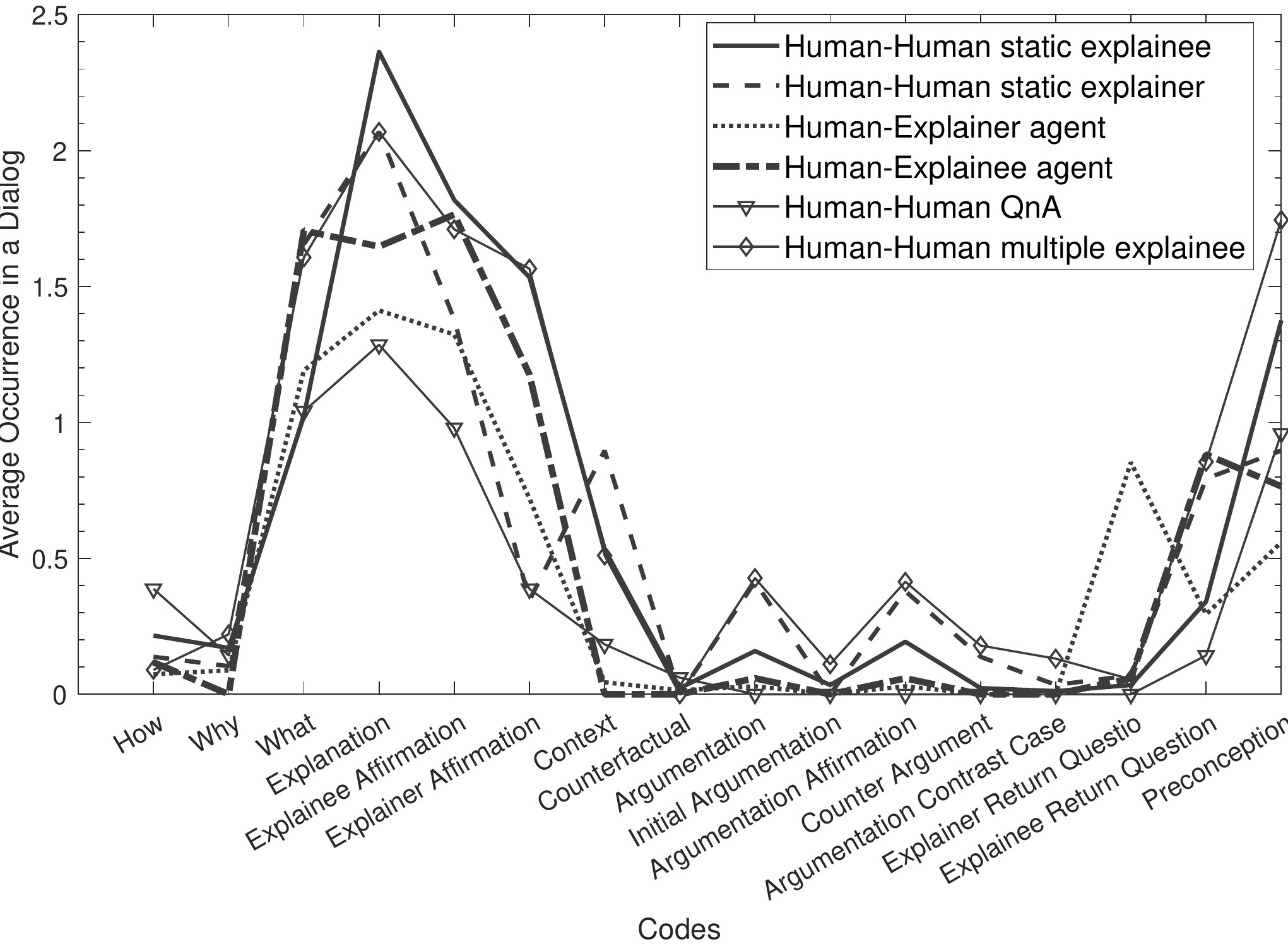}
\caption{Average code occurrence per dialogue in different explanation dialogue types} \label{fig2}
\end{figure}

The average code occurrence per dialogue in different dialogue types is depicted in Figure \ref{fig2}. In all dialogue types, a dialogue is most likely to have multiple \emph{what} questions, multiple \emph{explanations} and multiple \emph{affirmations}.

Argumentation is a key component of an explanation dialogue. The explainee can have different or contrasting views to the explainer regarding the explanation, at which point an argument can be put forth by the explainee. An argument in the form of an explanation that is not in response to a question can also occur at the very beginning of an explanation dialogue, where the argument set the premise for the rest of the dialogue. An argument is typically followed by an affirmation and may include a counter argument by the opposing party. From Figure \ref{fig2}, Human-Human dialogues with the exception of QnA have argumentation but Human-Agent dialogues lack any substantial occurrences of argumentation.

\subsubsection{Explanation Dialogue Termination Rule Analysis}

Participants should be able to identify when a dialogue ends. We analyse the different types of explanation dialogues to identify the codes that are most likely to signify termination.

\begin{figure}[!h]
\includegraphics[width=\columnwidth]{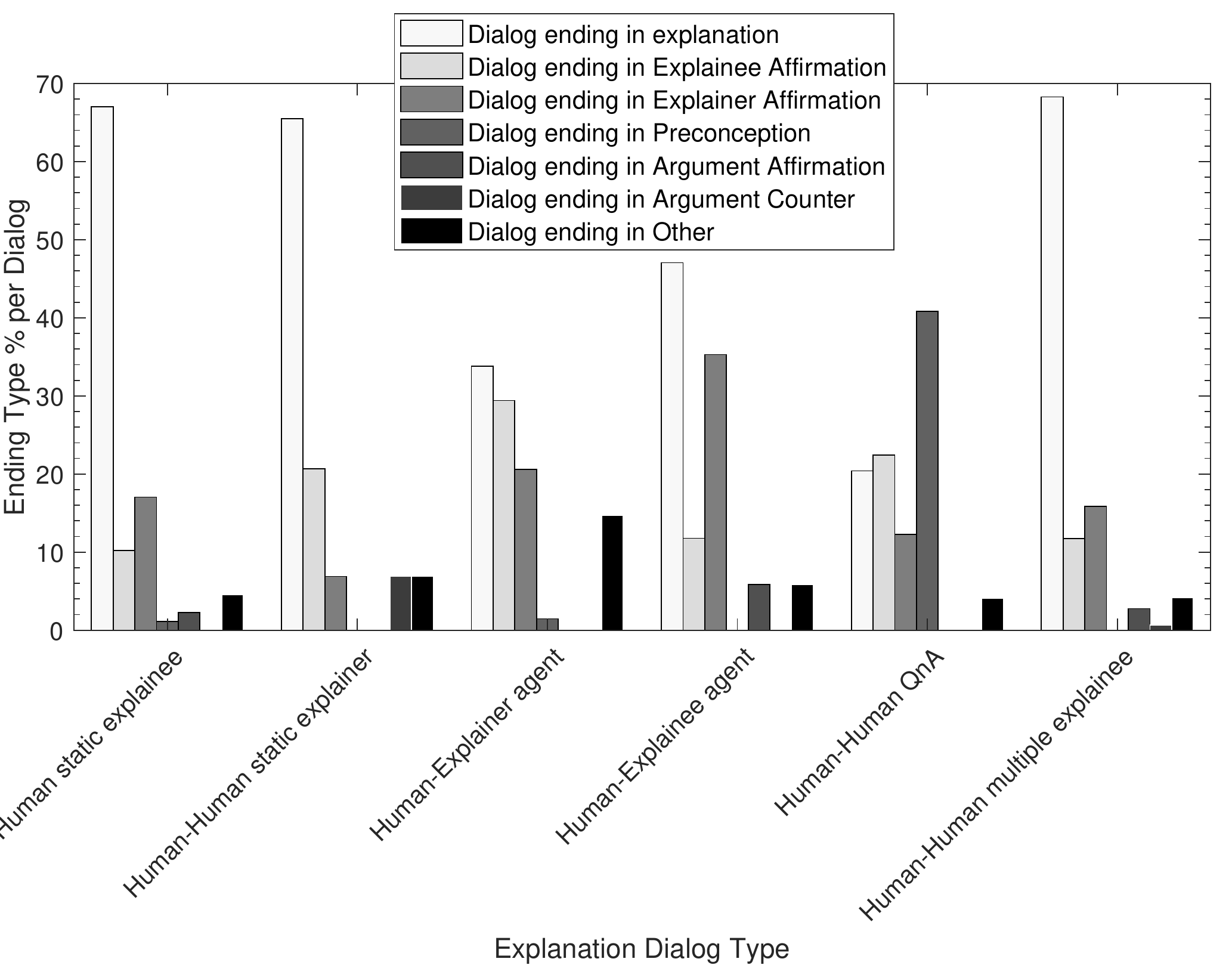}
\caption{Average code occurrence in per dialogue in different explanation dialogue types} \label{fig4}
\end{figure}

From Figure \ref{fig4}, all explanation dialogue types except Human-Human QnA type are most likely to end in an explanation. The second most likely code to end an explanation is explainer affirmation. Ending with other codes such as explainee and explainer return questions is presented by `Dialogue ending In Other' bar in Figure \ref{fig4}. It is important to note that although a dialogue is likely to end in an explanation, that dialogue can have previous explainee affirmations and explainer affirmations.

\subsection{Model Comparison}

\begin{figure}[!h]
\includegraphics[width=\columnwidth]{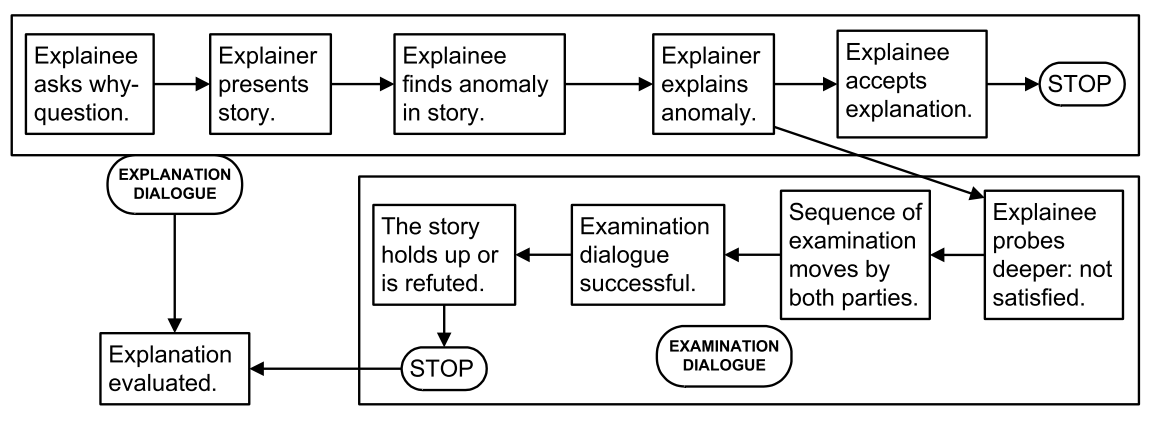}
\caption{Argumentation and explanation in dialogue~\cite{Reference8} } \label{fig6}
\end{figure}

We compare the explanation dialogue model, which also contains an argumentation sub-dialogue, by Walton~\cite{Reference8}. Walton proposed the model shown in Figure~\ref{fig6}, which consists of 10 components. This model focus on combining explanation and examination dialogues with argumentation.  A similar shift between explanation and argumentation/examination can be seen between our model and Walton's. According to the data sources, argumentation is a frequently present component of an explanation dialogue, which is depicted by the Explainee probing component in Walton's Model. The basic flow of explanation is the same between the two models, but the models differ in two key ways. First, is the lack of \emph{examination} dialogue shift in our model. Although we did not derive an examination dialogue, a similar shift of dialogue can be seen with respect to affirmation states. That is, our `examination' is simply the explainee affirming that they have understood the explanation.  Second is Walton's focus on the evaluation of the successfulness of an explanation in the form of examination dialogue, whereas our model focus on delivering an explanation in a natural sequence without an explicit form of explanation evaluation.

Thus, we can see similarities between Walton's conceptual model and our data-driven model. The differences between the two are at a more detailed level than at the high-level, and we attribute these differences to the grounded nature of our study. While Walton proposes an idealised model of explanation, we assert that our model captures the subtleties that would be required to build a natural dialogue for human-agent explanation.

\section{Empirical Validation}

In this section we discuss the validation of the derived explanation interaction protocol. We conducted a human-agent study in which an agent provides explanations using our model. The purpose of the study is to test whether the proposed model holds in a human-agent setting, and in particular, that the human participants follow the dialogue model when interacting with an artificial agent.

\subsection{Study}

\begin{figure}[!t]
\includegraphics[width=\columnwidth]{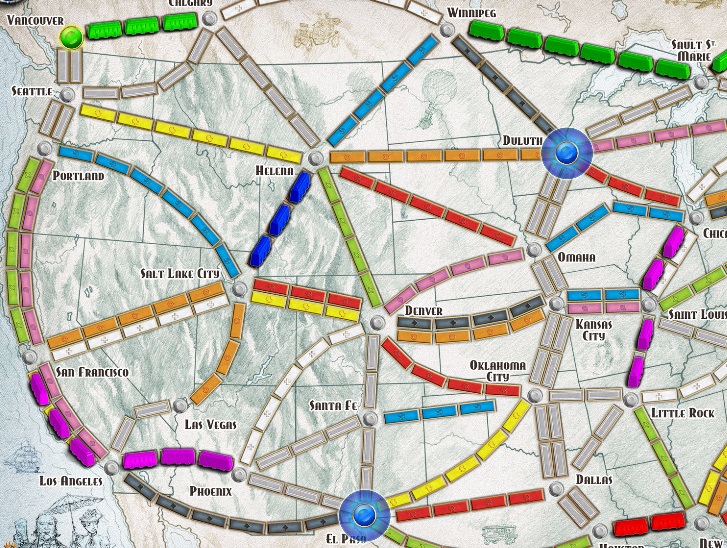}
\caption{Ticket to Ride Computer Game} \label{fig8}
\end{figure}

We conducted our study using the \textit{Ticket to Ride}\footnote{https://www.daysofwonder.com/tickettoride/en/} online computer game in a co-located competitive setting, previously used by~\citet{Newn} and~\citet{Singh}, in a university usability lab. The basic layout of the game is shown in Figure~\ref{fig8}. In this game, players must compete to build train routes between two cities, with each player building at least two such routes. For the purpose of the study, a game is played between two players who we term as the \textit{player} and the \textit{opponent}. The \textit{player} is assisted by an intelligent software agent that predicts the intentions and plans of the opponent. It is important to note that for a two player game, each route can be claimed only by one player. This allows players to block each other deliberately or otherwise, therefore inferring the intent of the opponent is beneficial for winning the game. We use the intent recognition algorithm of~\citet{Singh} to predict the opponent's future moves. The algorithm uses gaze data from an eye tracker and the actions of the opponent to formulate the possible plans (e.g. most probable routes the opponent can take). The opponent's gaze will also appear as a heat map on top of the player's Ticket to Ride game screen\footnote{Video capture of an experiment is provided in supplementary material at \url{https://explanationdialogs.azurewebsites.net/supp.zip}}. The agent communicates the predictions and their explanations to the player through a chat window.

To evaluate our model, we adopted a \textit{Wizard of OZ} approach described by~\citet{Dahlback}, meaning that the natural language generation of the explanation agent is played by a human `wizard', but this is unknown to the human participant. It is important to note that only the natural language generation is delegated to the wizard, while the agent generates and visualize plans (a set of connected train routes) in a separate interface in order to assist the wizard. Wizard has access to the visualized plans, most gazed at routes and most gazed at cities. The argument for using Wizard of Oz (WoZ) technique as opposed to a natural language implementation is twofold. First, to gather high quality empirical data related to the model bypassing the limitation that exist in natural language interfaces~\cite{Dahlback}. Second, having an interaction that closely resembles human discourse~\cite{Dahlback} allows the human to have more natural responses. Wizard of Oz techniques have been demonstratively shown to successfully evaluate conversational agents~\cite{chaves2018single, dubiel2018towards}, human-robot interactions ~\cite{Ball:2017:RAT:3183913.3183918} and automated vehicle-human interactions~\cite{mahadevan2018communicating, palmeiro2018interaction}  .

The \textit{Wizard} uses the prediction information of the agent and translates it to a more natural dialogue,  enabling us to get empirical data on natural explanation dialogues between player and the agent. The player is informed that he/she can communicate with the agent in a natural discourse. A prediction of the game includes a route that opponent might build (e.g from Pittsburgh to Houston) and a city area opponent might be interested in (e.g Interested around Houston). Predictions are generated from the implemented intent recognition algorithm of~\citet{Singh}. The wizard follows a simple natural language template: prediction followed by the explanation. The explanation template can include one or more of the following in any order: gaze explanation (e.g the opponent has been repeatedly gazing at that path) and causal history explanation (e.g the opponent has already built some paths along that route).

The protocol of the Wizard is outlined as follows. The Wizard follows the locutions, termination rules and combination rules of the dialogue model. Predictions and explanations of the agent is translated to natural language using the template described above by the wizard according to the nature of the locution used. Players (experiment participants) can ask questions and present argument in natural language. Players can initiate  dialogues as well as reply to the Wizard at any time in the game, and are in-control of the frequency of the interaction. The wizard follows a static failure response to any interactions that failed or unable to provide predictions and explanations (e.g I'm unable to answer that). Note that in the case of a participant using an invalid locution or a control dialogue, the dialogue will fail and wizard will end the dialogue with a termination rule.

Parameters of the experiments are as follows. In total, we obtained 101 explanation dialogues across 14 experiments. Players were from the same university, aged between 23 and 31 years ($M=27.2$). Players were observed through an observation room, in which the wizard was located. The duration of each experiment had an upper bound of 30 minutes ($M=20.15$), limited by the duration of the game-play, with the ability to end early if the game is won by either side before 30 minutes (game ends when all trains have been used by either side). During game-play the player has the freedom to engage in conversation with the agent or play the game disregarding the agent, thus each experiment yields a different number of dialogues ($M=7.21$). Conversations between player and the wizard were carried out using a chat client, through which we recorded the dialogue data. Extracted data was then analysed according to locutions, control dialogues and their sequences.

\subsection{Results and Findings}

\begin{figure}[!h]
\includegraphics[width=\columnwidth]{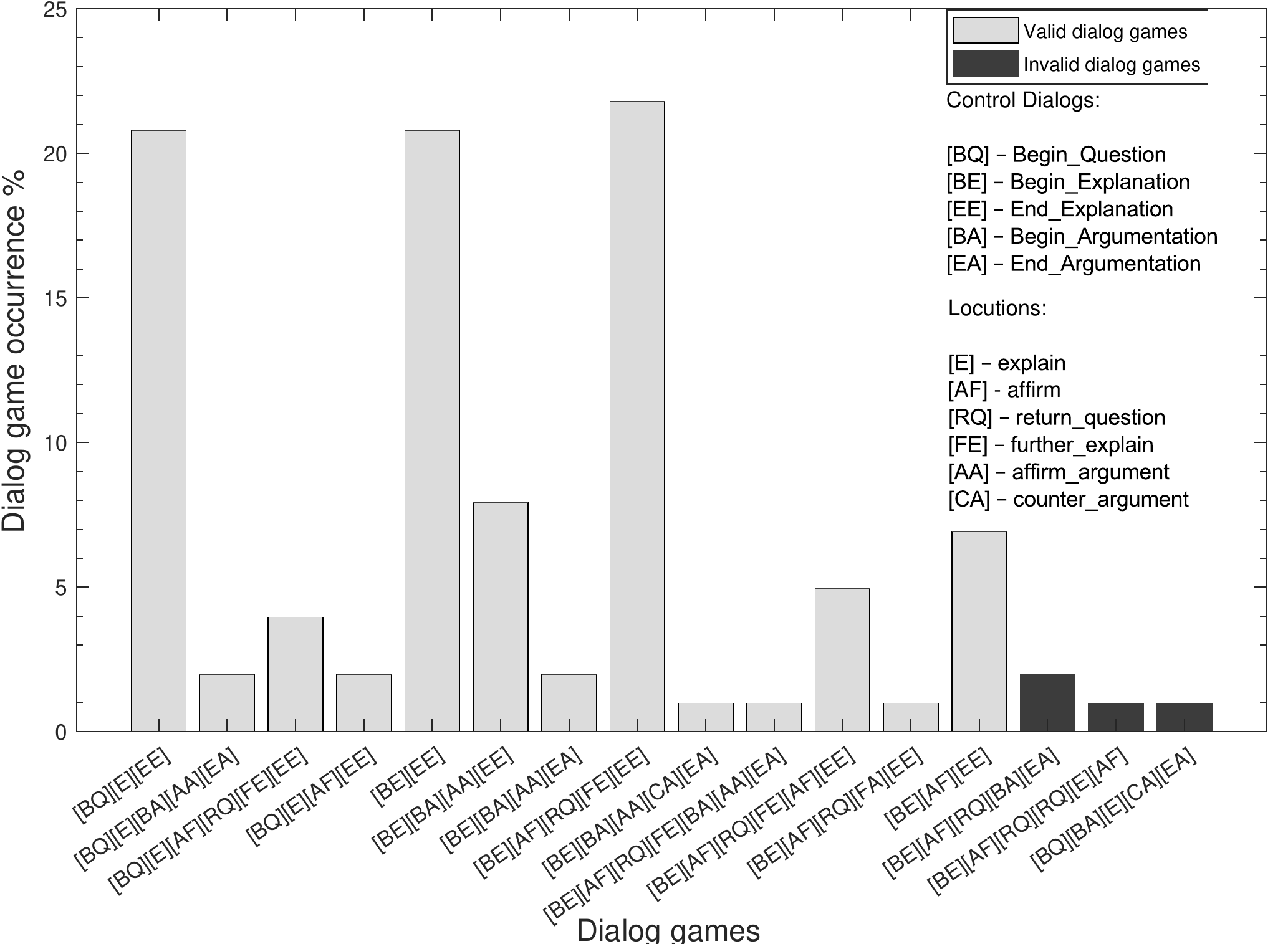}
\caption{Empirical results of human-agent dialogue games.} \label{fig7}
\end{figure}

Figure~\ref{fig7} illustrates the explanation dialogue games in the human-agent study. Control dialogues and locutions are depicted by shortened tags, which forms a sequence when combined. An example of a dialogue game using the tags would be [BQ][E][AF], which corresponds to [\textit{Begin\_question} $\Rightarrow$ \textit{explain} $\Rightarrow$ \textit{affirm}] in locutions and control dialogues. Figure \ref{fig7} shows percentages a specific dialogue game type occurred, and invalid dialogue games.

The proposed explanation dialogue model held true for 96 out of 101 dialogue game instances we observed. Figure \ref{fig7} indicates the 5 invalid dialogue game types that occurred. These dialogues became invalid dialogue games according to our model because of the parallelization of dialogue combination moves. For example, consider the dialogue game [BE][AF][RQ][BA][EA]. Here, after affirm and \textit{return\_question} locution, \textit{Begin\_Argumentaion} control dialogue occurs. This sequence is illegal according to the model. If parallelization is allowed, \textit{Begin\_Argumentaion} control dialogue can occur without waiting for a termination rule (e.g. affirm, explain). We attribute this limitation to the nature of the grounded data where parallelization cannot be accurately captured. This limitation can potentially be rectified by introducing parallelization \cite{McBurney2002} to the combination rules.

\section{Conclusion}

Explainable Artificial Intelligent systems can benefit from having a proper interaction protocol that can explain their actions and behaviours to the interacting users. Explanation naturally occurs as a continuous and iterative socio-cognitive process that involves two (sub)processes: a cognitive process and a social process. Most prior work is focused on providing explanations without sufficient attention to the needs of the explainee, which reduces the usefulness of the explanation to the end-user. 

In this paper, we propose a interaction protocol for the socio-cognitive process of explanation that is derived from different types of natural conversations between humans as well as humans and agents. We formalise the model using the Agent Dialogue Framework \cite{McBurney2002} as a new atomic dialogue type~\cite{Walton1995-WALCID} of explanation with an embedded argumentation dialogue, and we analyse  the frequency of occurrences of patterns. To empirically validate our model, we undertook a human behavioural experiment involving 14 participants and a total of 101 explanation dialogues. Results indicate that our explanation dialogue model can closely follow Human-Agent explanation dialogues. Main contribution of this paper lies in the formalized interaction protocol for explanation dialogues that is grounded on data, a secondary contribution is the coded (tagged) explanation dialogue data-set of 398 dialogues\footnote{coded (tagged) dialogue data-set is provided in supplementary material at \url{https://explanationdialogs.azurewebsites.net/supp.zip}}. By following a data-driven approach, proposed model captures the structure and the sequence of an explanation dialogue more accurately and allow natural interactions than explanations from existing models. The main contribution of this paper is a grounded interaction protocol derived from explanation dialogues, formalized as a new atomic dialogue type~\cite{Walton1995-WALCID} in the ADF. XAI systems that deal in explanation and trust will benefit from such a model in providing better, more intuitive and interactive explanations.

In future work, we aim introduce parallelism to the model with respect to locutions and combination rules as founded to be present in human-agent dialogues from the study. Further evaluation can be done by introducing other forms of interaction modes such as visual interactions which may introduce different forms of combination and termination rules.

\begin{acks}
The research described in this paper was supported by the University of Melbourne research scholarship (MRS); SocialNUI: Microsoft Research Centre for Social Natural User Interfaces at the University of Melbourne; and a Sponsored Research Collaboration grant from the Commonwealth of Australia Defence Science and Technology Group and the Defence Science Institute, an initiative of the State Government of Victoria. We also acknowledge the support of Joshua Newn and Ronal Singh,  given in conducting the study. 
\end{acks}


\bibliographystyle{ACM-Reference-Format}  
\balance  
\bibliography{main}  

\end{document}